\documentclass[conference]{IEEEtran} 
\usepackage{float}
\IEEEoverridecommandlockouts
\usepackage{booktabs}
\usepackage{pifont} 

\usepackage{cite}
\usepackage{amsmath,amssymb,amsfonts}
\usepackage{algorithmic}
\usepackage{graphicx}
\usepackage{textcomp}
\usepackage{xcolor}
\def\BibTeX{{\rm B\kern-.05em{\sc i\kern-.025em b}\kern-.08em
    T\kern-.1667em\lower.7ex\hbox{E}\kern-.125emX}}
\begin{document}

\title{TRACE-PCa: Predicting Prostate Cancer Progression from Longitudinal MRI \\ During Active Surveillance
}


\author{
\IEEEauthorblockN{
Hongye Zeng\textsuperscript{1},
Shreeram Athreya\textsuperscript{2},
Dingyuan Dai\textsuperscript{2},
Steve Raman\textsuperscript{1},\\
Leonard Marks\textsuperscript{4},
William Speier\textsuperscript{1,5},
Corey Arnold\textsuperscript{1,5,6,*}
}\\
\IEEEauthorblockA{
\textsuperscript{1}Department of Radiological Sciences,
\textsuperscript{2}Department of Electrical \& Computer Engineering,\\
\textsuperscript{4}Department of Urology,
\textsuperscript{5}Department of Bioengineering,
\textsuperscript{6}Departments of Pathology \& Laboratory Medicine\\
\textit{University of California, Los Angeles}, Los Angeles, USA
}
\thanks{\textsuperscript{*}Corresponding author: cwarnold@ucla.edu}
}


\maketitle

\begin{abstract}
Active surveillance (AS) is the preferred strategy for favorable-risk prostate cancer, yet current protocols rely on scheduled repeat biopsies, most of which reveal no progression and are unnecessary. Existing risk-stratification tools operate on single time-point imaging or depend on explicit lesion segmentation, limiting their ability to capture longitudinal change and excluding patients without an MRI-visible lesion. 
In this study, we propose an end-to-end temporal and multimodal model for predicting pathological progression during AS without lesion segmentation. We encode each serial scan with a pretrained 3D MRI foundation model and introduce a temporal attention gate that recalibrates the multi-visit features to amplify focal imaging changes associated with progression. The gated imaging representation is then fused with clinical variables in a multimodal framework to estimate the probability of progression.
Validated on a longitudinal AS cohort, our approach consistently outperforms competing baselines and performs comparably to the radiologist assessment representing current clinical practice. It maintains high negative predictive value while achieving higher positive predictive value, demonstrating its potential to safely reduce unnecessary biopsies during surveillance. Code will be released upon acceptance.
\end{abstract}


\section{Introduction}
Prostate cancer is the most common cancer and the second deadliest cancer among American men, with an estimated 299,010 new cases and 35,250 deaths in 2024 \cite{raychaudhuriProstate2025}. Active surveillance (AS) has emerged as the preferred management strategy for favorable-risk prostate cancer, aiming to avoid or delay curative treatment and its associated morbidities, including urinary incontinence, sexual dysfunction, and bowel dysfunction \cite{carlssonLongterm2020,hamdyFifteenYear2023}. Current AS protocols rely on serial prostate-specific antigen (PSA) measurements, digital rectal examinations, magnetic resonance imaging (MRI), and periodic repeat biopsies for disease monitoring \cite{Clinically}. However, 66–90\% of repeat biopsies in AS cohorts demonstrate no pathological progression \cite{drostCan2018,portenChanges2011,Gleason}, and many of these invasive procedures are therefore unnecessary under current uniform surveillance protocols. This high rate of uninformative biopsies imposes a substantial burden on patients. 
Personalized, non-invasive tools that can reliably identify patients with true disease progression and spare those who remain stable from unnecessary biopsy are therefore urgently needed \cite{mooreBest2023}.

\begin{figure}[t!]
    \centering
    \includegraphics[width=\columnwidth]{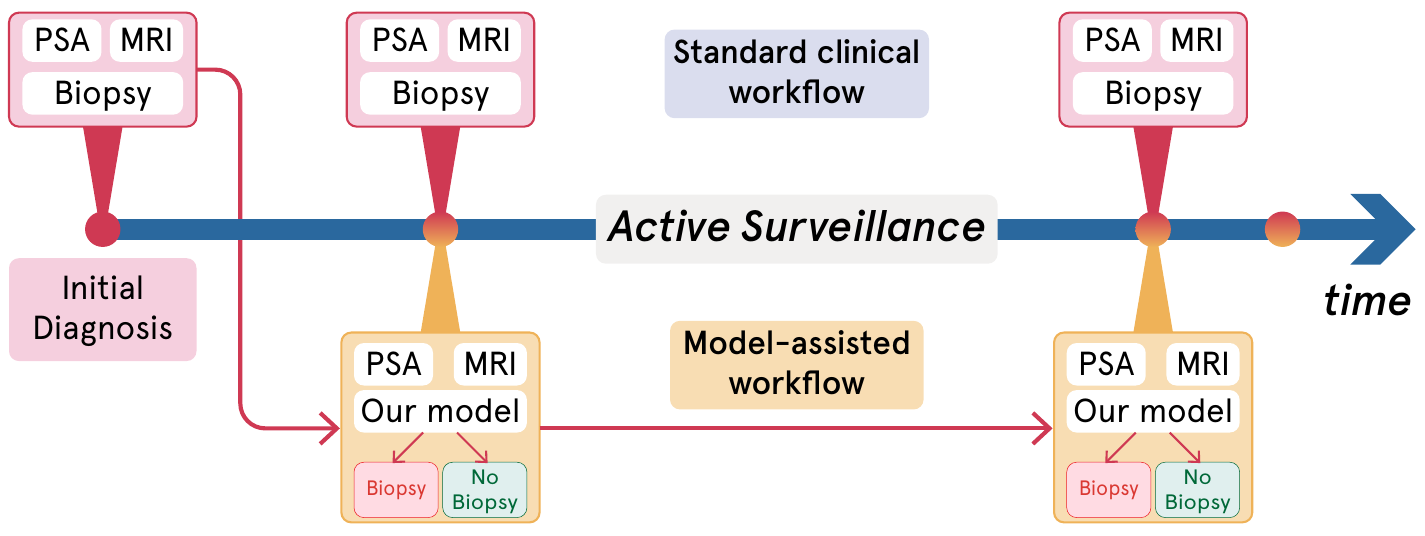}
    \caption{Standard AS monitoring workflow (top) versus the model-assisted workflow (bottom). In the standard protocol, all AS patients undergo scheduled repeat biopsy. The proposed model jointly analyzes baseline and follow-up bp-MRI scans with clinical variables to generate a patient-level risk score, enabling personalized biopsy decisions.}
    \label{fig:overview}
\end{figure}

Risk stratification during AS has historically relied on clinical nomograms incorporating PSA kinetics, Gleason grade, tumor volume, and Prostate Imaging Reporting and Data System (PI-RADS) scores derived from magnetic resonance imaging (MRI) \cite{leeDeveloping2022,cooperbergTailoring2020,tomerPersonalised2021,lightDevelopment2023,tomerShared2022}. Deep learning applied to prostate MRI offers a more reproducible assessment, and many models have shown considerable promise in automatically detecting and classifying clinically significant prostate cancer \cite{hammInteractive2023,leeMRIbased2025,shaoMRI2025,caiFully2024,fransenPatient2025,liProstAtlasDiff2025,paproskiPredicting2026,vasconcelosordonesNovel2025}. However, whether based on PI-RADS or these deep learning models, both operate on single time-point imaging and fail to leverage the longitudinal dynamics informative of progression during AS. While dedicated frameworks for scoring radiological change across serial scans have been proposed to address this \cite{englmanPRECISE2024,mooreReporting2017}, such assessments remain dependent on subjective visual interpretation.

The aforementioned limitations create a compelling clinical need for automated tools that can objectively quantify radiological change across serial MRI to better inform repeat biopsy decisions in AS. Longitudinal MRI analysis has been explored through delta radiomics pipelines, in which manually segmented lesions are used to extract and compare hand-engineered imaging features across timepoints to predict pathological progression \cite{midyaDelta2023,sushentsevTime2023,sushentsevComparative2022}. While these studies demonstrated superior predictive performance over single time-point approaches, their reliance on hand-engineered features and small cohorts limited generalizability.

Temporal deep learning frameworks, which jointly encode serial imaging data to learn disease trajectory directly without hand-crafted feature extraction, have demonstrated consistent success in other longitudinal medical imaging settings including Alzheimer's disease progression \cite{sunDeep2026, kimLearningbased2025a}, lung \cite{ardilaEndtoend2019,paezLongitudinal2023} and breast cancer screening \cite{donnellyAsymMirai2024,karamanLongitudinal2024, zhouDtMamba3D2026}, establishing the broader feasibility of this paradigm for learning clinically meaningful signals from serial scans. Within prostate AS specifically, a small number of studies have incorporated longitudinal MRI into deep learning pipelines \cite{roestAIassisted2023a,wangPredicting,wallaengenMpMRIBased2026,roestDevelopment2025}. However, these approaches have uniformly relied on explicit lesion segmentation as a necessary prerequisite for temporal analysis, with imaging features extracted from detected lesion regions and compared across serial scans. This dependency is a critical limitation, as lesion segmentation in prostate MRI is inherently challenging and error-prone \cite{liProstAtlasDiff2025,sahaArtificial2024}, particularly for the small and subtle lesions characteristic of favorable-risk cancer in the AS setting. 

In this study, we propose TRACE-PCa, a temporal and multimodal model that traces longitudinal radiological change from serial biparametric MRI (bp-MRI) scans and clinical variables to predict pathological progression during AS. Unlike prior longitudinal approaches that depend on explicit lesion segmentation, our model learns radiological change directly from serial whole-prostate MRI in an end-to-end manner, without any intermediate segmentation step. This eliminates the compounding error of lesion-dependent pipelines and allows the model to capture imaging dynamics across the entire gland, remaining applicable to all AS patients regardless of lesion visibility. To this end, we adopt a large-scale 3D MRI foundation model to extract image features from serial scans, mitigating the challenge of learning longitudinal representations from limited labeled data. We further propose a temporal attention gate that recalibrates the image features and selectively amplifies focal imaging changes associated with pathological progression. We validate our approach on a large longitudinal AS cohort, demonstrating consistent improvement over other baselines, confirming the value of explicitly modeling temporal dynamics for progression risk prediction.

\section{Methods}
\begin{figure*}[ht!]
    \centering
    \includegraphics[width=\textwidth]{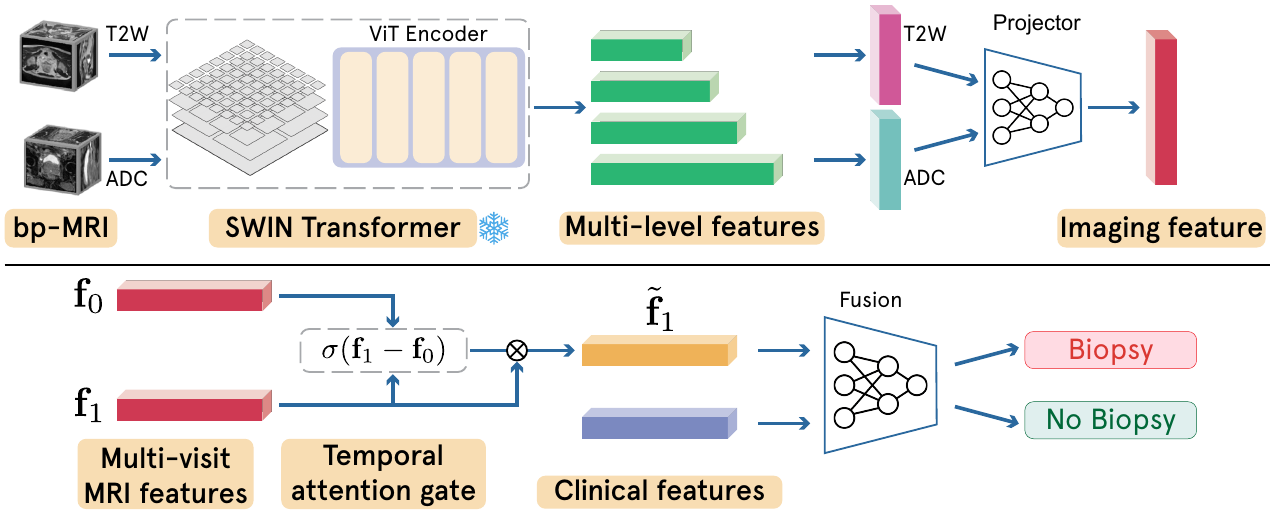}
    \caption{Overview of the proposed framework. \textbf{Top:} For each visit, the T2W and ADC volumes are encoded by a frozen Swin Transformer, producing image features that are pooled and projected into a single imaging feature. \textbf{Bottom:} The imaging features of the first ($\mathbf{f}_0$) and current ($\mathbf{f}_1$) visits are passed to the temporal attention gate, which recalibrates the current-visit feature into $\tilde{\mathbf{f}}_1$ to emphasize progression-relevant change. This temporal representation is fused with clinical features to predict pathological progression (biopsy vs.\ no biopsy).}
    \label{fig:method}
\end{figure*}

Figure~\ref{fig:method} presents an overview of our framework, which predicts biopsy-confirmed pathological progression from a pair of serial bp-MRI examinations and clinical variables. Given a baseline and a current visit, each examination is first encoded into a compact image representation by a 3D foundation model applied in a Siamese manner. The two visit representations are then compared by a temporal attention gate, which uses their difference to recalibrate the current-visit features and emphasize progression-relevant change. Finally, the resulting temporal image representation is fused with clinical variables and passed to a classifier to estimate the probability of pathological progression.

\subsection{MR image encoding}

Each surveillance examination provides an axial T2-weighted (T2W) volume and an apparent diffusion coefficient (ADC) volume, treated as two separate imaging inputs. Each volume is reoriented to a common coordinate system, center-cropped to $96\times96\times96$ voxels, and intensity-standardized within its nonzero region, yielding the preprocessed inputs $X_{\mathrm{T2W}}, X_{\mathrm{ADC}} \in \mathbb{R}^{96\times96\times96}$. We encode both volumes with Triad \cite{Wang2026}, a 3D vision foundation model that couples a Swin Transformer backbone with a SimMIM self-supervised objective to learn transferable volumetric MRI representations without task-specific labels. This design makes Triad well suited to prostate MRI and particularly advantageous in AS cohorts, where progression events are scarce. We denote the encoder by $\mathcal{T}$, initialize it with the publicly released Swin-Base weights, and keep it frozen, applying it in a Siamese manner to both modalities with shared parameters.


Applying $\mathcal{T}$ to each modality yields hierarchical feature maps $H_{\mathrm{T2W}}=\mathcal{T}(X_{\mathrm{T2W}})$ and $H_{\mathrm{ADC}}=\mathcal{T}(X_{\mathrm{ADC}})$, spanning four levels with channel dimensions $\{96,192,384,768\}$ and progressively coarser spatial resolution. We apply 3D global average pooling to each level, concatenate the pooled descriptors across both modalities, and map the result through a projector $\phi$ into a $d$-dimensional embedding. The full image-to-feature mapping is thus
\begin{equation}
f = \phi\big(\operatorname{GAP}(H_{\mathrm{T2W}}),\, \operatorname{GAP}(H_{\mathrm{ADC}})\big) \in \mathbb{R}^{d},
\end{equation}
where $\operatorname{GAP}(\cdot)$ denotes level-wise 3D global average pooling followed by concatenation across levels. Applying this mapping to the baseline and current examinations produces the paired representations $f_0, f_1 \in \mathbb{R}^{d}$ used for temporal modeling. All backbone features are extracted without gradient computation and cached, so that only the projector and temporal module are trained.

\subsection{Temporal attention gate}
Given the paired representations $f_0, f_1 \in \mathbb{R}^{d}$ of the baseline and current examinations, we characterize the radiological change between the two timepoints by an element-wise temporal difference that preserves the direction of change in each feature channel:
\begin{equation}
    \delta = f_1 - f_0.
\end{equation}

Rather than treating $\delta$ merely as an auxiliary feature, the temporal attention gate (TAG) uses it to recalibrate the last-visit representation. The difference is passed through a sigmoid function to produce a channel-wise gate $g \in (0,1)^{d}$, which then modulates $f_1$ via element-wise multiplication:
\begin{equation}
    g = \sigma(\delta), \qquad \tilde{f_1} = f_1 \odot g,
\end{equation}
where $\sigma(\cdot)$ is the sigmoid function and $\odot$ denotes the Hadamard product.  By operating on the signed difference, the gate is sensitive to the direction of temporal change, preserving channels that increase across visits while down-weighting those that remain stable or decrease. This directs the model toward feature dimensions exhibiting focal progression-associated change, while suppressing temporally static background signal.

\subsection{Longitudinal and multimodal fusion}
The gated representation $\tilde{f_1}$ encodes the last-visit imaging features recalibrated by their longitudinal change, thereby summarizing the temporal imaging trajectory in a single vector. To incorporate non-imaging information, the clinical variables $c \in \mathbb{R}^{n}$—including age, PSA kinetics, PSA density, baseline grade group, and biopsy core involvement—are embedded by a dedicated projector $\psi(\cdot)$ into a shared latent space. The projected clinical embedding is concatenated with the gated imaging representation and passed to a classification head $h(\cdot)$ to predict the probability of pathological progression:
\begin{equation}
    \hat{y} = h\big(\big[\,\tilde{f_1} \,\|\, \psi(c)\,\big]\big),
\end{equation}
where $\|$ denotes channel-wise concatenation. This late-fusion design allows the imaging and clinical branches to be encoded independently before being jointly optimized for progression prediction, enabling the model to leverage complementary signals from serial imaging dynamics and routinely collected clinical measurements.

\subsection{Loss Function}
We formulate pathological progression as a binary classification task and optimize the model with focal loss~\cite{lin2017focal}. Progression events are far less frequent than non-progression outcomes, making the cohort highly imbalanced. Under standard cross-entropy, the abundant easy majority-class samples dominate the gradient and dilute the signal from the rare progression cases. Focal loss addresses this by down-weighting confidently classified samples, concentrating training on the ambiguous and misclassified cases that matter most for progression detection.


For the $i$-th sample, let $s_i$ denote the model output logit and $y_i \in \{0,1\}$ the progression label. The predicted probability and the probability assigned to the ground-truth class are
\begin{equation}
p_i = \sigma(s_i),
\qquad
p_{t,i} = y_i p_i + (1-y_i)(1-p_i),
\end{equation}
where $\sigma(\cdot)$ is the sigmoid function. The focal loss for sample $i$ is
\begin{equation}
    \begin{aligned}
    \mathcal{L}_{i} &= -\,\alpha_{t,i}\,(1-p_{t,i})^{\gamma}\,\log p_{t,i}, \\
    \alpha_{t,i} &= \alpha\, y_i + (1-\alpha)(1-y_i),
    \end{aligned}
\end{equation}
where the modulating factor $(1-p_{t,i})^{\gamma}$ down-weights easy examples and the balancing term $\alpha_{t,i}$ adjusts the relative weight of positive and negative samples. We set $\alpha=0.8$ and $\gamma=2$. The training objective is the mean of $\mathcal{L}_{i}$ over each minibatch.

\section{Experiments and results}
\subsection{Dataset}
We utilize an AS cohort comprising 424 patients with favorable-risk prostate cancer who underwent serial bp-MRI examinations, resulting in 1,221 longitudinal scans. Repeat biopsy pathological outcomes, graded by fellowship-trained genitourinary pathologists according to the Gleason grade group (GG) system, served as ground truth. Pathological progression is defined as a transition from $GG\leq2$ at baseline to $GG>2$ at a subsequent visit, yielding 82 positive progression cases.

Table~\ref{tab:dataset_comparison} compares our dataset against existing longitudinal AS studies. Our cohort is substantially larger, with more total scans. Critically, it also includes patients without an MRI-visible lesion at baseline, whereas prior work has uniformly required a visible lesion as an inclusion criterion, excluding patients with radiologically occult disease. All prior approaches further rely on lesion segmentation, manual or automated, as a prerequisite for feature extraction. Our end-to-end framework requires no lesion-level annotation, enabling application across the full AS population regardless of lesion visibility. This reflects a more clinically representative and challenging setting than prior benchmarks.

\subsection{Implementation details and evaluation metrics}
Our model was trained using the AdamW optimizer with a learning rate of $1\times10^{-4}$ and weight decay of $1\times10^{-4}$, with a cosine annealing learning rate schedule over 60 epochs. A batch size of 8 was used, and Focal Loss ($\alpha=0.8$, $\gamma=2$) was employed to address class imbalance between progression and non-progression cases. Model selection was based on validation AUC after a 5-epoch warm-up period, with early stopping applied after 10 epochs without improvement. Patient-level five-fold cross-validation was used for all experiments, ensuring that all scan pairs from a given patient were assigned to the same fold. All experiments were conducted on a single NVIDIA Tesla V100 GPU. Model performance was evaluated using Accuracy, Specificity, F1 score, negative predictive value (NPV), positive predictive value (PPV), and area under the receiver operating characteristic curve (ROC-AUC), averaged across all five folds.

\subsection{Performance comparison with other methods}
We compare our proposed method against three model-based baselines and the radiologist assessment using PI-RADS, which represents current clinical practice. The clinical nomogram uses only clinical variables with an XGBoost classifier. The lesion-based and gland-based baselines use delta-radiomics features and combine the same clinical variables for prediction. The required lesion and gland segmentations were obtained using segmentation models trained on the public PI-CAI dataset \cite{sahaArtificial2024}. All learned methods are evaluated under the same 5-fold cross-validation protocol.

As shown in Table~\ref{tab:comparison}, our model achieved the strongest overall discrimination, with a mean ROC-AUC of 0.704 (0.028), substantially outperforming the clinical nomogram (0.579), the lesion-based model (0.618), and the gland-based model (0.635). The monotonic improvement from the clinical-only nomogram to the lesion-based and gland-based models—all of which share the same clinical variables—indicates that adding imaging features improves discrimination, and that expanding the imaging context from restricted lesion regions to the whole prostate is further beneficial. Our temporal framework, which additionally models change across serial scans, extends this trend and achieves the largest margin. Beyond AUC, our model attained the highest accuracy (0.708), specificity (0.716), F1 (0.476), and PPV (0.392) among all methods. These gains are particularly relevant in AS, where most repeat biopsies reveal no progression, so that higher specificity translates directly into fewer unnecessary biopsies, while the higher PPV means a positive prediction is more likely to reflect true progression. The lesion- and gland-based baselines had low specificity (0.471 and 0.503), meaning they wrongly flagged many stable patients as progressing. Their slightly higher NPV (0.932 and 0.923 vs.\ 0.903) largely reflects the class imbalance in the cohort, and a model that predicts the majority class for nearly all cases attains high NPV with little genuine discriminative ability.

\begin{figure}[t]
    \centering
    \includegraphics[width=\linewidth]{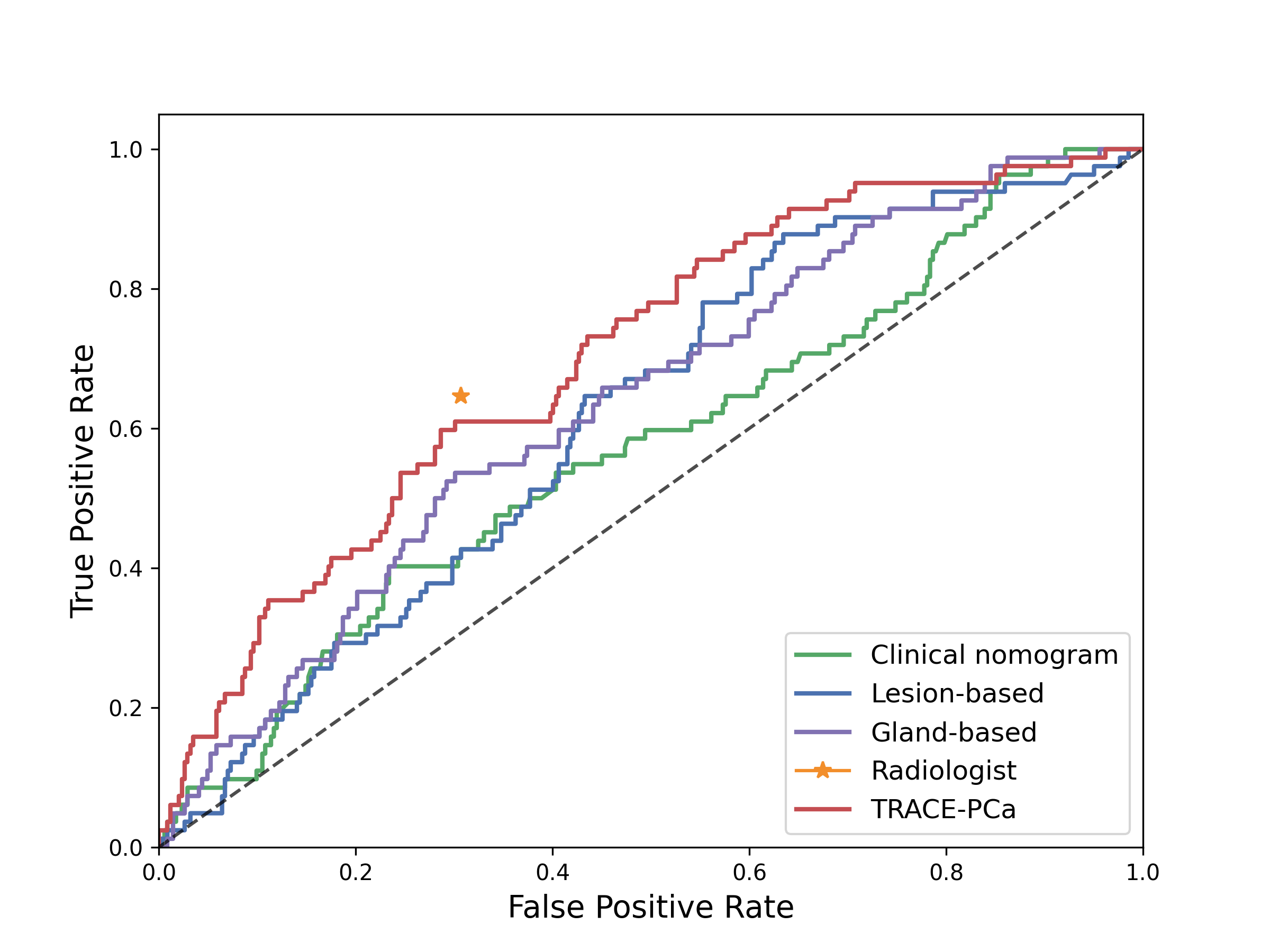}
    \caption{ROC curves comparing our model with baseline methods. The radiologist assessment (PI-RADS $>$ 3) is shown as a single operating point (orange star).}
    \label{fig:roc}
\end{figure}

Figure~\ref{fig:roc} shows the corresponding ROC curves. Our model consistently dominates the model-based baselines across most of the operating range, with the largest margin at low false positive rates (FPR$<$0.4), the clinically relevant region for minimizing unnecessary biopsies while retaining sensitivity to true progression. The clinical nomogram shows the weakest discrimination, approaching the diagonal at low FPR. The radiologist operating point (orange star) lies slightly above our ROC curve, achieving a marginally higher true positive rate at its fixed operating threshold. Unlike this single fixed point, however, our model produces a continuous risk score, allowing the decision threshold to be tuned to the desired sensitivity–specificity trade-off for different clinical scenarios. Together, these results indicate that our end-to-end temporal framework captures progression-relevant imaging change more effectively than models relying on single-timepoint or lesion-restricted representations.

\begin{table}[t]
    \centering
    \caption{Comparison of datasets used in longitudinal prostate MRI studies for AS progression prediction.}
    \label{tab:dataset_comparison}
    \begin{tabular}{lcccc}
        \toprule
        Dataset & Positive & Total & No-lesion & Lesion \\
         & /total patients & scans & positive & Segmentation\\
        \midrule
        Cleveland \cite{midyaDelta2023} & 28/50 & 80  & \ding{55} & Manual \\
        Cambridge \cite{sushentsevTime2023} & 28/76 & 297  & \ding{55} & Manual \\
        Miami \cite{wangPredicting} & 38/206 &  458 & \ding{55} & Automated \\
        Ours & 82/424 & 1221 & \ding{51} & Not required \\
        \bottomrule
    \end{tabular}
\end{table}

\begin{table*}[h]
    \centering
    \caption{Performance comparison with baseline methods. Results are reported as mean (std) over 5-fold cross-validation.}
    \label{tab:comparison}
    \begin{tabular}{lcccccc}
    \toprule
    Method & Accuracy & Specificity & F1 & NPV & PPV & ROC-AUC \\
    \midrule
    Clinical nomogram & 0.620 (0.125) & 0.623 (0.195) & 0.388 (0.051) & 0.875 (0.025) & 0.303 (0.095) & 0.579 (0.072) \\
    Lesion-based & 0.540 (0.106) & 0.471 (0.166) & 0.413 (0.025) & 0.932 (0.041) & 0.280 (0.034) & 0.618 (0.027) \\
    Gland-based & 0.562 (0.123) & 0.503 (0.182) & 0.417 (0.056) & 0.923 (0.038) & 0.288 (0.046) & 0.635 (0.083) \\
    Radiologist & 0.684 (0.040) & 0.693 (0.040) & 0.441 (0.060) & 0.892 (0.035) & 0.336 (0.044) & -- \\
    TRACE-PCa & 0.708 (0.090) & 0.716 (0.141) & 0.476 (0.032) & 0.903 (0.020) & 0.392 (0.097) & 0.704 (0.028) \\
    \bottomrule
    \end{tabular}
\end{table*}

\begin{table*}[ht]
    \centering
    \caption{Ablation on multimodal fusion and longitudinal modeling. Results are reported as mean (std) over 5-fold cross-validation.}
    \label{tab:ablation_inputs}
    \begin{tabular}{lcccccc}
    \toprule
    Input & Accuracy & Specificity & F1 & NPV & PPV & ROC-AUC \\
    \midrule
    Image-only & 0.585 (0.080) & 0.552 (0.109) & 0.405 (0.033) & 0.890 (0.010) & 0.284 (0.037) & 0.623 (0.043) \\
    Clinical-only & 0.686 (0.113) & 0.719 (0.201) & 0.372 (0.157) & 0.880 (0.044) & 0.315 (0.085) & 0.594 (0.126) \\
    Multimodal & 0.693 (0.052) & 0.716 (0.098) & 0.425 (0.042) & 0.885 (0.029) & 0.339 (0.026) & 0.657 (0.057) \\
    Longitudinal+multimodal & 0.708 (0.090) & 0.716 (0.141) & 0.476 (0.032) & 0.903 (0.020) & 0.392 (0.097) & 0.704 (0.028) \\
    \bottomrule
    \end{tabular}
\end{table*}

\subsection{Ablation study}

\subsubsection{Temporal attention gate}
Table~\ref{tab:ablation_components} compares model performance with and without the temporal attention gate (TAG). While AUC improves only marginally with TAG (0.704 vs. 0.702), Specificity increases substantially from 0.541 to 0.716 (0.175), and PPV improves from 0.313 to 0.392 (0.079). This indicates that while both variants achieve similar overall discriminative ranking, TAG produces substantially fewer false positives at the operating threshold, directly reducing the number of patients who would be recommended for unnecessary biopsy. Despite the marginal change in AUC, this improvement in specificity and PPV is clinically meaningful for reducing unnecessary biopsies in AS.

\subsubsection{Last-visit features}
As shown in Table~\ref{tab:ablation_components}, we compare a variant using only the temporal difference feature $\delta$ against our full model, which additionally incorporates the last-visit representation $f_1$. Incorporating $f_1$ improves AUC from 0.660 to 0.704 and Specificity from 0.663 to 0.716, indicating that the current-visit representation provides complementary information beyond the temporal change signal alone, consistent with the clinical intuition that both the trajectory of change and the most recent disease state are informative for progression risk assessment.

\subsubsection{Contribution of multimodal fusion and longitudinal modeling}
Table~\ref{tab:ablation_inputs} reports an ablation over the input components of our model, isolating the contributions of multimodal fusion and longitudinal modeling. The first three variants all operate on single time-point imaging and/or clinical data, while the last additionally models change across serial scans. Using imaging or clinical variables alone yields limited discrimination (AUC 0.623 and 0.594, respectively). Although the clinical-only variant attains relatively high accuracy and specificity, this reflects a bias toward predicting the majority stable class rather than genuine discriminative ability, as evidenced by its low AUC and F1. Combining the two modalities improves AUC to 0.657 and raises F1 and PPV over either single-modality variant, indicating that imaging and clinical information are complementary. Introducing longitudinal modeling on top of multimodal fusion (Longitudinal+multimodal, our full model) provides a further substantial gain, increasing AUC from 0.657 to 0.704 and improving F1 (0.425$\rightarrow$0.476) and PPV (0.339$\rightarrow$0.392) while maintaining specificity. This confirms that explicitly modeling temporal change across serial scans contributes information beyond what static multimodal features capture, and is the largest single contributor to the final performance.

\subsubsection{Image encoder feature stage} 
Figure~\ref{fig:ablation_hyper}a shows model performance when using features extracted from different backbone stages. Performance decreases monotonically as deeper stages are used, from an AUC of 0.704 at stage 1 to 0.539 at stage 4. This indicates that low-level features, which retain fine-grained spatial detail, are more informative for capturing subtle longitudinal imaging change than high-level semantic features, which are increasingly abstracted away from voxel-level detail as network depth increases. Based on this result, we adopt stage-1 features for our final model.

\subsubsection{Projector dimension} 
\begin{figure}[t]
    \centering
    \includegraphics[width=1\linewidth]{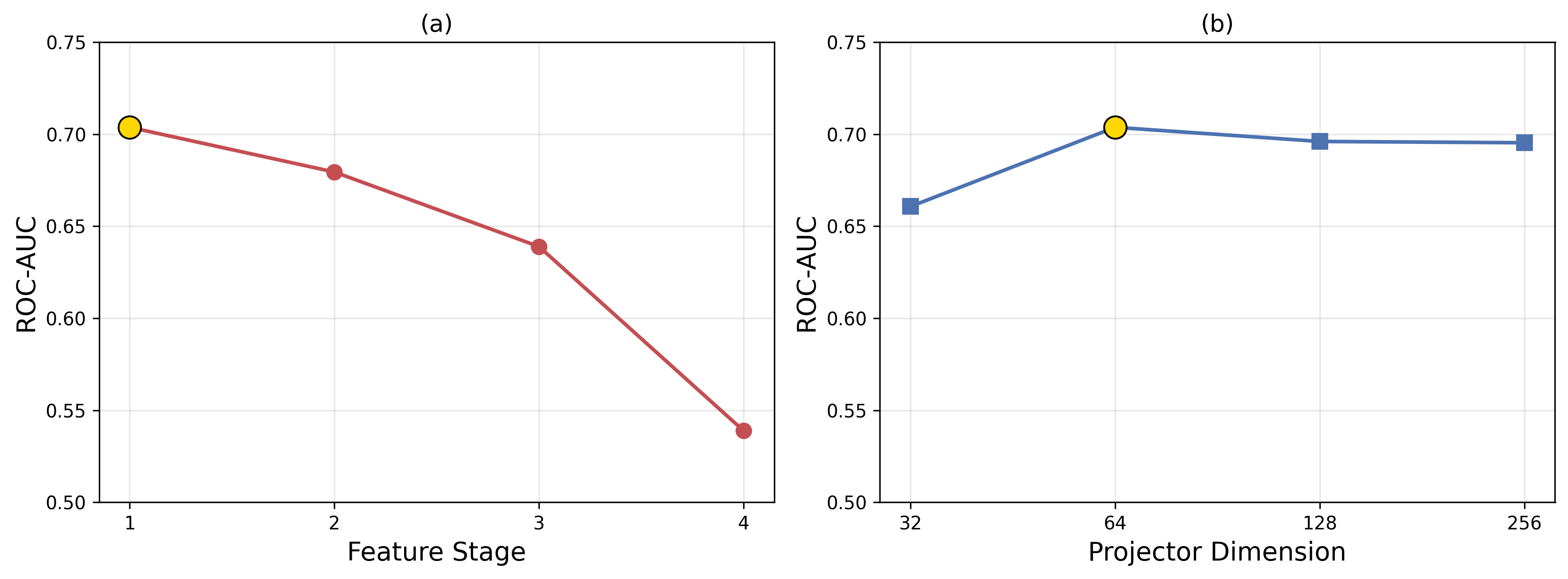}
    \caption{Ablation on model hyperparameters. (a) Effect of the image encoder feature stage. (b) Effect of the projector output dimension. The peak in each panel is marked with a yellow point.}
    \label{fig:ablation_hyper}
\end{figure}

Figure~\ref{fig:ablation_hyper}b shows the effect of the projector output dimension. Performance is relatively stable across dimensions (AUC 0.661--0.704), with the best result achieved at a dimension of 64. Smaller (32) and larger (128, 256) dimensions yield modest but consistent performance drops, suggesting that a moderate projector dimension provides sufficient capacity for temporal feature encoding without introducing unnecessary parameters that may increase overfitting risk given the limited size of the longitudinal cohort. We select a projector dimension of 64 for our final model.

\begin{table}[t]
    \centering
    \caption{Ablation study on temporal model components. Values are reported as mean (STD).}
    \label{tab:ablation_components}
    \begin{tabular}{cccccc}
    \toprule
    $\delta$ & $f_1$ & TAG & AUC & PPV & Specificity \\
    \midrule
    \checkmark & &  & 0.660 (0.014) & 0.375 (0.102) & 0.663 (0.247) \\
    \checkmark & \checkmark &  & 0.702 (0.036) & 0.313 (0.035) & 0.541 (0.157) \\
    \checkmark & \checkmark & \checkmark & 0.704 (0.028) & 0.392 (0.097) & 0.716 (0.141) \\
    \bottomrule
    \end{tabular}
\end{table}

\section{Conclusion}
In this study, we proposed TRACE-PCa, an end-to-end temporal deep learning framework for predicting pathological progression in prostate cancer active surveillance from longitudinal bp-MRI and clinical variables. By operating directly on the whole prostate without requiring lesion segmentation, our method is applicable across the full AS population, including patients without an MRI-visible lesion at baseline, a group systematically excluded by prior lesion-dependent approaches. Central to our framework is a temporal attention gate that recalibrates the current-visit representation using the encoded temporal change, improving specificity and PPV over naive temporal fusion while preserving discriminative performance. Across a large longitudinal cohort, our model consistently outperformed clinical, lesion-based, gland-based, and PI-RADS-based baselines on most metrics. While our cohort is larger than those in prior studies, it remains single-institutional and moderate in size; future work will extend validation to larger, multi-institutional cohorts.

\bibliographystyle{IEEEtran}
\bibliography{refs}

\end{document}